%% file: main.tex
\documentclass[runningheads]{llncs}

\pdfoutput=1

\input{header}
\usepackage[T1]{fontenc}
\usepackage{graphicx}
\begin{document}
\title{Navigating Data Scarcity using Foundation Models: A Benchmark of Few-Shot and Zero-Shot Learning Approaches in Medical Imaging}
\titlerunning{Navigating Data Scarcity using Foundation Models in Medical Imaging}
%
\author{Stefano Woerner\inst{1} \and
Christian F. Baumgartner\inst{1,2}
}
\authorrunning{S. Woerner and C.F. Baumgartner}
%
\institute{Cluster of Excellence \enquote{Machine Learning}, University of Tübingen, Germany\\
\email{\{firstname.lastname\}@uni-tuebingen.de}
\and
Faculty of Health Sciences and Medicine, University of Lucerne, Switzerland
}
\maketitle              
\begin{abstract}
Data scarcity is a major limiting factor for applying modern machine learning techniques to clinical tasks.
Although sufficient data exists for some well-studied medical tasks, there remains a long tail of clinically relevant tasks with poor data availability.
Recently, numerous foundation models have demonstrated high suitability for few-shot learning (FSL) and zero-shot learning (ZSL), potentially making them more accessible to practitioners.
However, it remains unclear which foundation model performs best on FSL medical image analysis tasks and what the optimal methods are for learning from limited data.
We conducted a comprehensive benchmark study of ZSL and FSL using 16 pretrained foundation models on 19 diverse medical imaging datasets.
Our results indicate that BiomedCLIP, a model pretrained exclusively on medical data, performs best on average for very small training set sizes, while very large CLIP models pretrained on LAION-2B perform best with slightly more training samples.
However, simply fine-tuning a ResNet-18 pretrained on ImageNet performs similarly with more than five training examples per class.
Our findings also highlight the need for further research on foundation models specifically tailored for medical applications and the collection of more datasets to train these models.

\end{abstract}
\section{Introduction}
Machine learning is revolutionizing the field of medical imaging and diagnostics, offering capabilities that were previously unattainable.
However, these advancements typically depend on the availability of large, well-annotated datasets.
For many medical applications, such as the diagnosis of rare diseases, collecting these types of datasets is often infeasible.
Consequently, in many real-world scenarios, there is often insufficient data to effectively train highly performant deep learning models. Additionally, computational resources are frequently limited, which poses further challenges in training or even fine-tuning state-of-the-art models.

Few-shot learning (FSL) has shown great potential in addressing these data-scarce applications.
With effective FSL strategies, clinics and medical researchers could potentially train models using their own small datasets and achieve performance levels acceptable for clinical practice.
Few-shot learning is most commonly performed through fine-tuning of a large pretrained model on the smaller, domain-specific, target dataset.
Recently, several large models, known as foundation models, have been published after being trained on vast amounts of data~\cite{pmlr-v139-radford21a, Cherti_2023, oquab2023dinov2, zhang2023biomedclip}.
Many such models have been shown to have excellent generalization capabilities, and to be highly suitable for FSL.
However, no large-scale studies exist which compare FSL performance of different pretrained models across a broad and diverse array of medical imaging domains.
A number of foundation models are also capable of zero-shot learning (ZSL) by searching for the highest correspondence between the representations of the input image and a language prompt.
Similarly, there are no works rigorously comparing the ZSL capabilities of different foundation models on a diverse range of medical tasks. 

In this paper, we present the first large-scale study comparing the FSL and ZSL performance of various publicly available pretrained models across a diverse set of medical imaging domains.
We conduct our study on the recently released MedIMeta dataset~\cite{woerner2024comprehensive}, which is comprised of 19 different datasets from 10 different imaging modalities and anatomical regions. 
In comprehensive experiments we evaluate 16 publicly available models that have been pretrained on different medical and non-medical data sources.
Because fine-tuning very large models is not practical within the computational budget of most clinicians and researchers, we limited ourselves to exploring strategies that are possible to perform in the realistic scenario of having access to a single mid-range to high-end GPU.
Within these constraints we explore a linear probing strategy as well as fine-tuning.
For the five models in our benchmark that support ZSL, we also benchmark their ZSL capabilities with different prompt styles.

Our experiments yield a number of practical insights and actionable recommendations.
We make code to reproduce our results and adapt our experiments publicly available.\footnote{\url{https://github.com/StefanoWoerner/medimeta-fsl-benchmark}}

\section{Methods}

\subsection{Dataset}
To allow us to study the FSL and ZSL performance on wide array of different image modalities and tasks, we conduct our experiments on the recently released MedIMeta dataset~\cite{woerner2024comprehensive, stefano_woerner_2024_7884735}.
MedIMeta is a highly standardized meta-dataset compiled from 19 publicly available datasets, and covering 10 different imaging modalities.
We use the main (i.e. first) task for each of the 19 datasets.
We refer the reader to~\cite{woerner2024comprehensive, stefano_woerner_2024_7884735} for detailed descriptions of the sub-datasets and tasks. 

\subsection{Simulation of FSL and ZSL tasks}

We artificially construct multiple FSL tasks from each of MedIMeta's datasets by randomly sampling $n$ labeled training samples and $10$ unlabeled query samples per class from each dataset.
We ensure that no images of the same subject are spread over the two sets. We coin these individual FSL tasks a \textit{task instance}.
To ensure robust FSL performance measurement, we randomly generate 100 task instances for each dataset and average the results.
In order to investigate the effect of increasing numbers of labeled training samples we repeat all experiments for $n\in\{1,2,3,5,7,10,15,20,25,30\}$.
In addition we also simulate task instances with $n=0$, i.e. only query samples, for the ZSL evaluation. 

\subsection{Pretrained Models}\label{sec:models}

We evaluate three distinct pretraining paradigms: supervised pretraining, self-supervised pretraining, and contrastive language-image pretraining (CLIP). In the following we briefly describe the specific architectures and pretraining data.

\boldparagraph{Fully Supervised Models.}
We investigate the widely used \textbf{Residual Networks (ResNet) architecture}~\cite{He_2016_CVPR} in the variations ResNet18, ResNet50, and Resnet101, all of which have been pretrained on the ImageNet dataset~\cite{ILSVRC15}.

We further investigate the \textbf{Vision Transformer (ViT) architecture}~\cite{dosovitskiy2020image}.
Due to it's excellent performance on many computer vision benchmarks, the ViT has become a standard architecture and the basis of a large amount of further work.
We compare the base (ViT-B), large (ViT-L), and huge (ViT-H) architecture variations with patch sizes 16, 16 and 14, respectively.
We consider models pretrained on ImageNet~\cite{ILSVRC15} and on ImageNet21k~\cite{ridnik2021imagenet21k}.

\boldparagraph{Self-supervised Models.}
In this category we consider the \textbf{self-DIstillation with NO labels (DINO)} model \cite{caron2021emerging}.
We specifically focus on the recently released DINOv2 model~\cite{oquab2023dinov2} which relies on a ViT architecture that was pretrained using a self-supervised knowledge distillation approach.
The model was trained using a very large unlabeled but curated dataset assembled from various computer vision datasets.
The DINOv2 representations have been shown to be highly transferable across computer vision tasks~\cite{oquab2023dinov2}.
We consider the ViT-B, ViT-L, and giant (ViT-g) variations with patch size 14.

\boldparagraph{Contrastive Language-Image Pretraining.}
Lastly, we consider two CLIP models which employ contrastive learning to align images and text into a shared embedding space~\cite{pmlr-v139-radford21a}.

Firstly, we use the original \textbf{CLIP model} with the weights for ViT-B and ViT-L provided by OpenAI~\cite{pmlr-v139-radford21a}.
These models have been pretrained on 400 million image-text pairs collected from the internet.
Although the specific composition of this dataset is not traceable, it is likely that a small portion of medical data was included.
In addition to its unique ZSL capabilities, the CLIP model was also shown to perform extremely well on computer vision FSL tasks by training a linear probe on the final image-encoder representations~\cite{pmlr-v139-radford21a}.

Secondly, we use the ViT-H and ViT-g models trained on LAION-2B~\cite{schuhmann2022laion5b} provided by OpenCLIP~\cite{Cherti_2023}, an open source reimplementation of OpenAI's CLIP.
LAION-2B contains 2 billion image-text pairs extracted from common crawl~\cite{commoncrawl} and is the English language subset of the larger LAION-5B~\cite{schuhmann2022laion5b} dataset.
Similar to the OpenAI data, the inclusion of small amounts of medical data is likely.

We also investigate the \textbf{BiomedCLIP model}~\cite{zhang2023biomedclip} which uses the same ViT architecture as the base version of the original CLIP, but replaces the text encoder with PubMedBERT~\cite{Gu_2021}, a language model tailored for the biomedical domain.
BiomedCLIP was pretrained on 15 million text-image pairs extracted from PubMed articles (PMC-15M). This is the only model in our study that was trained exclusively on medical data. BiomedCLIP can be employed for FSL and prompt-based ZSL in the same manner as CLIP. 

\subsection{Few-shot Learning Strategies}

We evaluate two model adaptation strategies: fine-tuning and linear probing. 

\textbf{Fine-tuning} involves initializing a network with pretrained weights, and then continuing the training of all weights in the network with an FSL task instance.
The last linear layer (classification layer) is replaced with a new layer matching the number of classes in the target task.
For most foundation models, which commonly have hundreds of millions or even billions of parameters, fine-tuning is computationally infeasible for many practitioners.
We therefore only evaluate the fine-tuning strategy on the ResNet-18 and ResNet-50 variants. 

Similarly, \textbf{linear probing} involves initializing a network with pretrained weights, and attaching a new classification layer.
However, in linear probing the backbone network is frozen, and a simple linear classifier is trained on the final representations of the network.
This was shown to lead to strong FSL performance assuming the base network is able to extract useful image features~\cite{pmlr-v139-radford21a}.
Since only a linear classifier is trained on the image features produced by the pretrained network, this strategy is computationally much cheaper than fine-tuning the complete network, making it feasible to use with large foundation models.

We conduct an extensive \textbf{hyper-parameter search} on a separate set of sampled FSL tasks for both fine-tuning and linear probing.
For each of the models and each number of labeled samples $n$, we test two optimizers (SGD and Adam), two different head initialization strategies (Kaiming initialization~\cite{He_2016_CVPR}, initialization with all zeros~\cite{woerner2022strategies}), a range of learning rates between $10^{-5}$ and $0.1$, and a range of training steps between 5 and 200.
We evaluate all models from Section~\ref{sec:models} using their respective optimal parameters from the hyper-parameter search.

\subsection{Zero-shot learning strategy}

The CLIP~\cite{pmlr-v139-radford21a} and BiomedCLIP~\cite{zhang2023biomedclip} models have the capability of solving classification tasks with no labeled training examples by searching for the highest similarity between an input image and several text prompts corresponding to different target classes.
We test three different prompt templates.
First, we investigate simply using the class names extracted from the MedIMeta task definitions as prompts.
Secondly, we test two templates which add information about the imaging modality: \enquote{A \{domain\_identifier\} image where the \{task\_name\} is \{class\_name\}}, and \enquote{This \{domain\_identifier\} image shows [a] \{class\_name\}}.
All variables above are extracted from the MedIMeta task description files.
However, some class names and domain identifiers needed to be adjusted in order to form a grammatically correct and semantically meaningful sentences.

\subsection{Metrics}

We evaluate the performance for each dataset and each training set size $n$ using the area under the receiver operator curve (AUROC) averaged over all 100 task instances.
To obtain a measure of average performance across all datasets, we use the harmonic mean of the AUROCs from each dataset.

\begin{figure}[t!]
    \includegraphics{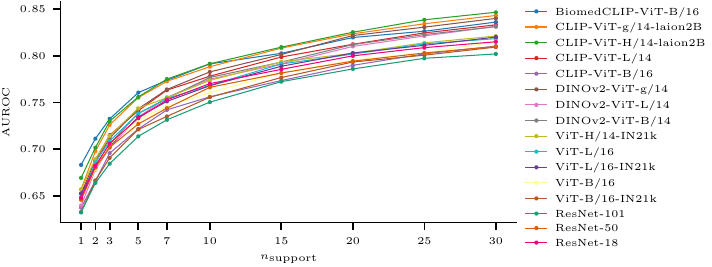}
    \caption{Harmonic mean AUROC over all 19 MedIMeta datasets}
    \label{fig:hm-auroc}
\end{figure}

\section{Experiments and Results}

We performed all FSL and ZSL experiments using all models and learning strategies as described above.
In the following, we describe our main findings.
All results can be found in Table~A.1 in the Supplementary Material.

\boldparagraph{The optimal hyperparameters were similar for all models.}
For all models the best-performing optimizer was Adam~\cite{kingma2017adam}. 
Further, initializing the classification head with zeros performed better or on par compared to Kaiming initialization~\cite{He_2016_CVPR}, in line with the findings in \cite{woerner2022strategies}.
For most models and $n$, using a learning rate of $10^{-4}$ with at least 120 training steps was optimal or close to optimal.

\begin{figure}[ht!]
    \centering
    \includegraphics{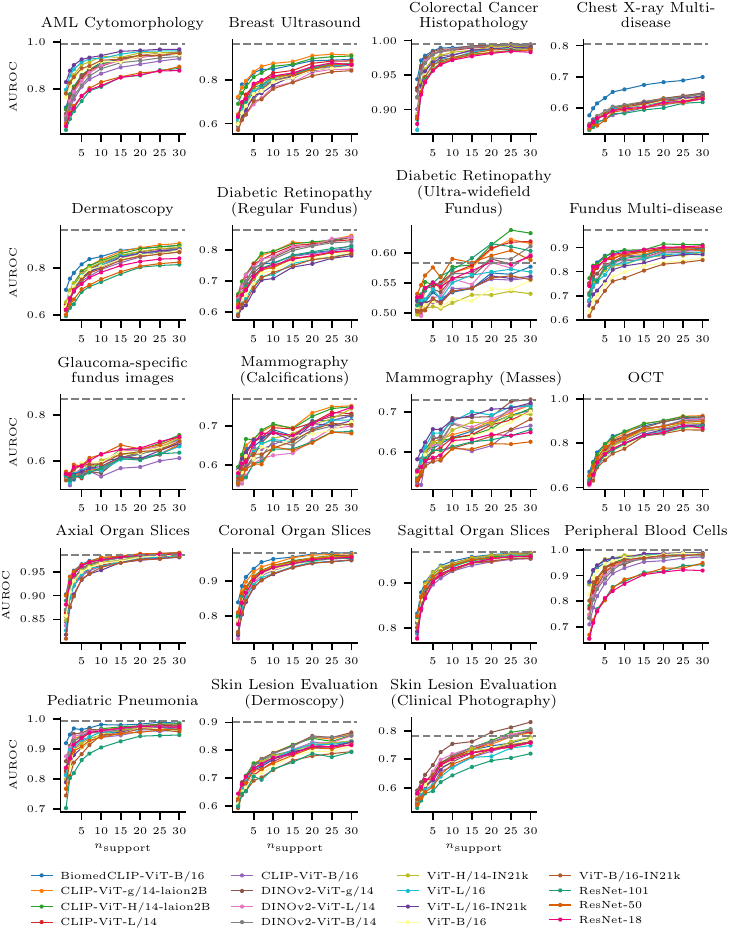}
    \caption{AUROC on the different datasets with fully supervised baseline from \cite{woerner2024comprehensive}. The fully-supervised performance is indicated by the black dotted line.}
    \label{fig:per-dataset-auroc}
\end{figure}

\boldparagraph{Linear probing with BiomedCLIP and CLIP-ViT-H yielded the best results on average.}
As can be seen in Fig.~\ref{fig:hm-auroc} CLIP-ViT-H on average outperformed other pretrained models for $n\geq 7$.
Interestingly, the performance of the ``huge'' variant of CLIP was slightly better than the even larger ``giant (g)'' variant.
However, for smaller $n$, BiomedCLIP, which was the only foundation model in our comparison trained entirely with medical data, outperformed its larger CLIP counterparts and performed on par with CLIP-ViT-H up to $n=10$.
We note that CLIP-based pretraining led to the best performance overall, underscoring the potential of contrastive language-image pretraining for FSL.

\boldparagraph{Linear probing performance on individual datasets was mixed.} 
The performance on the individual datasets shown in Fig.~\ref{fig:per-dataset-auroc} was mixed.
Interestingly, the method that performed best on average for $n\geq 7$ (CLIP-ViT-H) rarely performed the best on the individual datasets.
Rather it was consistently among the top-few approaches on most datasets leading to its high average performance.
BiomedCLIP, on the other hand, performed very well on some datasets (e.g. Chest X-ray Multi-disease, Dermatoscopy, and Pediatric Pneumonia), but more poorly on others (e.g. Ultra-widefield Fundus, or Mammography (Masses)).
We hypothesize that BiomedCLIP performed more strongly on images that were overrepresented in the PMC-15M pretraining dataset.
We conclude that in practice linear probing on the BiomedCLIP model might often be a good first attempt when working with very few labeled images, but it does not obviate thorough evaluation on a held-out test set.
With more training data, linear probing CLIP-ViT-H is likely the better option, especially since it does not display as much variablility throughout different imaging modalities as BiomedCLIP.

\begin{figure}[t!]
    \centering
    \includegraphics{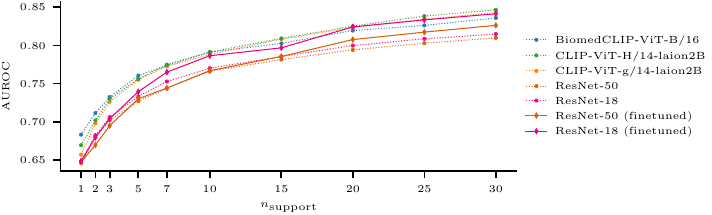}
    \caption{Harmonic mean AUROC across all 19 MedIMeta datasets of fine-tuned ResNet models with the best-performing linear probe as a point of comparison.}
    \label{fig:hm-acc-auroc-ft}
\end{figure}

\boldparagraph{FSL performed close to fully supervised learning for some tasks.} 
In Fig.~\ref{fig:per-dataset-auroc} we additionally show the fully supervised baseline performance on the official data splits reported in \cite{stefano_woerner_2024_7884735}.
We observed that for some tasks the 30-shot performance almost matched the fully supervised performance.
Indeed, for the Ultra-widefield Fundus dataset the FSL performance was substantially better than the fully supervised performance.
We believe this is due to the small number of training images in the official split of this dataset.
Nevertheless, this suggests that for some problems linear probing of a foundation model may be a better alternative than training a model from scratch with a small dataset.

\boldparagraph{Linear probes on large models beat fine-tuning of small models.}
In Fig.~\ref{fig:hm-acc-auroc-ft} it can be seen that linear probing with CLIP-ViT-H on average outperformed fine-tuning of the ResNet-18 and ResNet-50 for all $n$.
However, with more training data fine-tuning the ResNet-18 performed \emph{almost} as well as CLIP-ViT-H, and for $n\geq20$ the fine-tuned ResNet-18 outperformed BiomedCLIP.
Interestingly, fine-tuning ResNet-18 clearly outperformed fine-tuning ResNet-50, suggesting that a lower network complexity may be preferable in the FSL scenario.
Our findings suggest that while linear probing very large foundation models such as the CLIP-ViT-H on average may lead to small performance gains, the commonly used strategy of fine-tuning a ResNet-18 also performs strongly given sufficient data.
We note that while fine-tuning of foundation models may lead to even better results, this is computationally prohibitive for the majority of practitioners.

\boldparagraph{ZSL performance could not match FSL performance.}
The ZSL approaches had average AUROC scores ranging from $0.316$ to $0.397$, far below those of the 1-shot performance reported in Fig.~\ref{fig:hm-auroc} where average AUROCs range from $0.632$ to $0.683$.
This contradicts the findings of Radford et al.~\cite{pmlr-v139-radford21a} who showed that on computer vision tasks the CLIP model can often outperform linear probes in the ZSL setting.
We conclude that ZSL may not yet be a suitable strategy for general medical image analysis tasks. We report the ZSL results in Fig.~A.1 in the Supplementary Materials.

\begin{figure}[t!]
    \centering
	\includegraphics{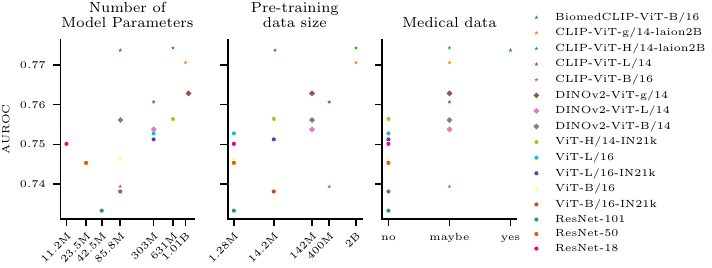}
    \caption{Model properties plotted against mean few-shot performance over the tested $n$. 
    The symbol indicates the type of pretraining (Supervised, DINO, CLIP).}
    \label{fig:model-properties}
\end{figure}

\boldparagraph{Model complexity and pretraining data size correlate with performance.}
In Fig.~\ref{fig:model-properties}, we explore the relation of the following three model properties to their linear probing performance: the model size, the number of samples in the pretraining data, and the type of pretraining data. 
We observed that there was a strong positive correlation between model size and few-shot performance as well as pretraining set size and few-shot performance.
While the non-medical CLIP-ViT-H, and CLIP-ViT-g clearly outperformed all other non-medical approaches on average, BiomedCLIP, which was trained on medical data exclusively, performed very well despite much smaller number of parameters and pretraining data size.
This underscores the need for building training sets which contain a diverse set of medical images
and for training advanced medicine-focused foundation models.

\section{Conclusion}
We performed the first large-scale study comparing the FSL and ZSL performance of a wide array of pretrained models on a diverse set of medical imaging data.
We found that, on average, in the very low data regime of $n\leq 5$ samples per class, a linear probe on BiomedCLIP was the best strategy.
However, with more data, linear probing of the CLIP-ViT-H model performed slightly better. 
While fine-tuning a ResNet-18 on average performed worse compared to a linear probe on CLIP-ViT-H, it still reached a high performance for $n\geq 20$.
We also observed a large variance between the performance on the different datasets emphasizing the need for cautious application of these technologies.
Our investigation further revealed that parameter-rich foundation models trained on very large non-medical datasets have very good FSL performance on medical tasks.
However, the strong performance of BiomedCLIP model on some datasets underscores the potential of foundation models specific to medical applications.

\begin{credits}
\subsubsection{\ackname} Funded by the Deutsche Forschungsgemeinschaft (DFG, German Research Foundation)
under Germany’s Excellence Strategy – EXC number 2064/1 – Project number 390727645.
The authors thank the International Max Planck Research School for Intelligent Systems
(IMPRS-IS) for supporting Stefano Woerner.

\subsubsection{\discintname} The authors have no competing interests to declare.
\end{credits}

\printbibliography

\clearpage
\input{supplementary}

\end{document}

%% file: header.tex
\usepackage[T1]{fontenc}
\usepackage[utf8]{inputenc}

\usepackage[maxbibnames=9, maxcitenames=2, backend=biber, natbib]{biblatex}

\usepackage[english]{babel}

\usepackage{csquotes}

\usepackage{amsmath}
\usepackage{amssymb}

\usepackage{isomath}

\usepackage{booktabs}
\usepackage{array}

\usepackage{algorithm, algpseudocode}

\usepackage{graphicx}
\graphicspath{{img/}}

\usepackage{pdfpages}

\usepackage{tikz}
\usetikzlibrary{positioning,shapes.geometric,fit}
\usepackage{pgfplots}
\pgfplotsset{compat=1.16}

\usepackage{siunitx}
\newcolumntype{U}{S[table-format=2.1(1), separate-uncertainty=true]}
\usepackage{multirow}

\usepackage[
  babel=true, 
  expansion=alltext,
  protrusion=alltext-nott, 
  final 
]{microtype}
\DisableLigatures{encoding = T1, family = tt* }

\newcommand{\boldparagraph}[1]{\vspace{.5\baselineskip plus 2pt}\noindent{\bf #1}}

%% file: supplementary.tex
\appendix
\renewcommand\thefigure{A.\arabic{figure}} 
\renewcommand\thetable{A.\arabic{table}}
\setcounter{figure}{0}
\setcounter{table}{0}

\section{Supplementary Materials}

\begin{figure}
	\centering
	\includegraphics{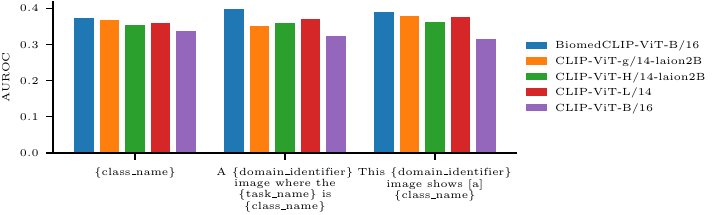}
	\caption{Harmonic mean AUROC across all 19 datasets with three zero-shot templates.}
	\label{fig:zero-shot-auroc}
\end{figure}

\vspace{-1\baselineskip}

\begin{table}[]
    \centering
    \caption{AUROC in \% for all MedIMeta datasets and all $n$.}
    \label{tab:my_label}
\end{table}

\vspace{-2\baselineskip}

\centering
\scalebox{.33}{

\begin{tabular}{lSSSSSSSSSSSSSSSSSSSS}
\toprule
{} & \multicolumn{20}{c}{$n=1$} \\
{} & {Harmonic Mean} & {aml} & {bus} & {crc} & {cxr} & {derm} & {dr\_regular} & {dr\_uwf} & {fundus} & {glaucoma} & {mammo\_calc} & {mammo\_mass} & {oct} & {organs\_axial} & {organs\_coronal} & {organs\_sagittal} & {pbc} & {pneumonia} & {skinl\_derm} & {skinl\_photo} \\

\midrule
BiomedCLIP-ViT-B/16 & \bfseries 68.3 & 72.0 & 71.9 & \bfseries 94.4 & \bfseries 57.6 & \bfseries 70.8 & 65.1 & 52.2 & 75.1 & 54.2 & 57.5 & 53.3 & \bfseries 67.0 & \bfseries 90.3 & \bfseries 84.0 & \bfseries 83.2 & 86.5 & \bfseries 92.0 & 62.1 & 57.8 \\
CLIP-ViT-g/14-laion2B & 65.7 & 68.1 & \bfseries 72.3 & 91.7 & 53.7 & 63.7 & 65.6 & 51.4 & 74.2 & 53.9 & 57.3 & 53.1 & 62.2 & 86.8 & 80.2 & 80.2 & 74.9 & 82.1 & 62.5 & 54.4 \\
CLIP-ViT-H/14-laion2B & 66.9 & 71.1 & 69.1 & 93.1 & 54.3 & 64.5 & \bfseries 65.8 & 52.5 & \bfseries 77.3 & 55.1 & \bfseries 59.4 & 55.3 & 65.0 & 87.0 & 79.7 & 81.4 & 78.2 & 83.4 & 62.3 & 55.9 \\
CLIP-ViT-L/14 & 65.3 & 68.8 & 63.7 & 93.1 & 54.1 & 64.4 & 65.0 & 50.6 & 75.2 & 53.6 & 56.9 & 51.6 & 62.7 & 86.3 & 81.1 & 81.1 & 77.1 & 82.8 & 60.2 & 54.8 \\
CLIP-ViT-B/16 & 63.7 & 66.5 & 62.6 & 90.1 & 52.9 & 62.6 & 60.9 & 51.1 & 67.4 & 53.5 & 55.5 & 52.3 & 61.5 & 84.4 & 78.5 & 78.6 & 71.0 & 81.6 & 61.6 & 54.7 \\
DINOv2-ViT-g/14 & 64.7 & 69.3 & 58.8 & 92.8 & 54.1 & 63.1 & 62.6 & 50.4 & 71.1 & 52.6 & 55.9 & 55.4 & 61.3 & 85.6 & 75.5 & 79.0 & 73.4 & 86.0 & 62.1 & \bfseries 59.1 \\
DINOv2-ViT-L/14 & 64.0 & 68.2 & 57.4 & 91.8 & 54.1 & 62.2 & 64.3 & 50.1 & 67.5 & 54.7 & 54.7 & 53.8 & 61.0 & 83.6 & 73.6 & 77.9 & 74.5 & 87.6 & 59.9 & 56.8 \\
DINOv2-ViT-B/14 & 64.7 & 67.4 & 59.1 & 91.8 & 54.8 & 63.6 & 63.7 & 52.3 & 67.7 & 54.7 & 55.5 & 53.4 & 61.9 & 86.9 & 78.5 & 81.1 & 73.5 & 83.7 & 60.6 & 56.4 \\
ViT-H/14-IN21k & 65.7 & 78.3 & 60.5 & 89.3 & 54.2 & 65.4 & 63.2 & 49.7 & 70.2 & 53.8 & 57.1 & 54.8 & 63.6 & 84.8 & 78.1 & 81.9 & 86.5 & 80.0 & 62.7 & 55.7 \\
ViT-L/16 & 64.8 & 79.6 & 61.6 & 87.0 & 54.2 & 63.1 & 59.4 & 52.1 & 68.3 & 53.7 & 57.1 & 54.3 & 61.9 & 82.7 & 75.0 & 78.9 & 87.3 & 81.3 & 60.4 & 53.8 \\
ViT-L/16-IN21k & 65.3 & \bfseries 82.8 & 59.7 & 89.2 & 54.0 & 62.2 & 58.7 & 51.9 & 65.9 & 54.4 & 58.6 & \bfseries 58.2 & 62.8 & 81.8 & 74.6 & 79.2 & \bfseries 87.6 & 79.0 & 61.2 & 56.1 \\
ViT-B/16 & 64.7 & 76.4 & 59.8 & 89.3 & 53.8 & 63.5 & 60.6 & 51.0 & 63.6 & 51.8 & 58.3 & 54.3 & 63.5 & 85.7 & 78.9 & 81.7 & 82.8 & 76.8 & 60.9 & 56.4 \\
ViT-B/16-IN21k & 63.3 & 77.9 & 57.1 & 89.0 & 54.3 & 61.9 & 59.0 & 50.2 & 61.6 & 51.6 & 56.5 & 52.7 & 61.7 & 80.9 & 75.2 & 80.0 & 80.2 & 74.6 & 59.3 & 56.0 \\
ResNet-101 & 63.2 & 62.3 & 63.0 & 88.6 & 52.9 & 59.6 & 63.6 & 51.3 & 70.4 & 54.0 & 55.0 & 53.4 & 63.4 & 88.9 & 80.3 & 82.4 & 65.5 & 70.4 & 59.9 & 52.9 \\
ResNet-50 & 64.6 & 65.1 & 64.3 & 88.7 & 53.3 & 60.1 & 62.3 & \bfseries 53.2 & 74.6 & 55.5 & 55.2 & 53.1 & 65.4 & 90.0 & 81.0 & 82.6 & 66.9 & 76.8 & 62.4 & 54.0 \\
ResNet-18 & 64.8 & 63.9 & 63.1 & 87.9 & 54.1 & 62.3 & 61.5 & 52.6 & 74.1 & 54.5 & 57.8 & 55.1 & 61.6 & 88.1 & 77.7 & 77.6 & 65.4 & 83.8 & \bfseries 64.5 & 58.1 \\
ResNet-50 (finetuned) & 64.7 & 64.5 & 63.8 & 87.3 & 50.7 & 60.0 & 62.3 & 51.8 & 71.8 & \bfseries 56.6 & 56.6 & 54.4 & 65.1 & 89.6 & 80.6 & 81.9 & 69.6 & 83.2 & 62.9 & 55.4 \\
ResNet-18 (finetuned) & 64.9 & 63.9 & 64.7 & 86.0 & 53.8 & 61.4 & 63.8 & 53.1 & 74.4 & 55.1 & 56.7 & 54.3 & 61.6 & 86.8 & 76.6 & 75.5 & 68.3 & 88.9 & 63.0 & 57.6 \\
\bottomrule
\end{tabular}

}
\\
\centering
\scalebox{.33}{

\begin{tabular}{lSSSSSSSSSSSSSSSSSSSS}
\toprule
{} & \multicolumn{20}{c}{$n=2$} \\
{} & {Harmonic Mean} & {aml} & {bus} & {crc} & {cxr} & {derm} & {dr\_regular} & {dr\_uwf} & {fundus} & {glaucoma} & {mammo\_calc} & {mammo\_mass} & {oct} & {organs\_axial} & {organs\_coronal} & {organs\_sagittal} & {pbc} & {pneumonia} & {skinl\_derm} & {skinl\_photo} \\

\midrule
BiomedCLIP-ViT-B/16 & \bfseries 71.1 & 77.6 & \bfseries 78.0 & \bfseries 97.1 & \bfseries 59.9 & \bfseries 75.5 & 68.3 & 51.2 & 80.0 & 52.5 & 58.8 & 55.5 & \bfseries 71.5 & \bfseries 94.0 & \bfseries 88.6 & \bfseries 87.9 & 91.7 & \bfseries 94.8 & 67.9 & 58.9 \\
CLIP-ViT-g/14-laion2B & 69.8 & 75.1 & 76.5 & 95.7 & 56.1 & 70.0 & \bfseries 70.0 & 50.2 & 80.5 & 52.6 & 62.1 & 56.5 & 67.7 & 92.6 & 87.0 & 86.7 & 82.3 & 90.3 & 67.7 & 57.8 \\
CLIP-ViT-H/14-laion2B & 70.2 & 77.3 & 74.0 & 96.0 & 55.7 & 70.3 & 69.8 & 52.8 & \bfseries 82.4 & 51.4 & \bfseries 62.6 & 55.9 & 71.0 & 91.7 & 85.8 & 86.5 & 85.6 & 89.6 & 66.9 & 58.8 \\
CLIP-ViT-L/14 & 68.8 & 75.4 & 69.7 & 95.9 & 55.9 & 68.8 & 68.1 & 52.7 & 79.7 & 53.1 & 59.6 & 54.0 & 67.4 & 90.4 & 85.9 & 85.6 & 83.8 & 87.3 & 64.8 & 57.8 \\
CLIP-ViT-B/16 & 66.6 & 71.8 & 66.8 & 93.9 & 54.2 & 67.1 & 65.8 & 50.8 & 74.4 & 49.6 & 58.6 & 51.7 & 66.3 & 89.5 & 83.8 & 83.8 & 77.5 & 86.0 & 65.5 & 56.9 \\
DINOv2-ViT-g/14 & 68.6 & 76.5 & 64.2 & 96.3 & 56.2 & 69.7 & 65.8 & 49.6 & 78.1 & 51.7 & 56.2 & 58.3 & 66.8 & 91.5 & 83.7 & 86.2 & 82.2 & 91.1 & 68.6 & \bfseries 61.9 \\
DINOv2-ViT-L/14 & 67.8 & 74.7 & 62.6 & 95.4 & 55.7 & 68.3 & 68.4 & 49.5 & 74.5 & 53.4 & 55.1 & 57.0 & 66.5 & 90.1 & 82.0 & 84.4 & 81.9 & 90.5 & 66.4 & 60.0 \\
DINOv2-ViT-B/14 & 68.1 & 73.1 & 64.8 & 95.3 & 56.3 & 68.7 & 66.2 & 51.8 & 73.7 & 52.6 & 57.2 & 58.2 & 66.2 & 91.6 & 84.5 & 85.6 & 80.7 & 87.2 & 66.3 & 58.6 \\
ViT-H/14-IN21k & 68.9 & 82.9 & 65.5 & 93.9 & 55.7 & 70.6 & 66.1 & 51.4 & 75.5 & 51.7 & 59.9 & 55.9 & 67.3 & 90.3 & 84.6 & 87.1 & 91.7 & 86.0 & 66.9 & 58.9 \\
ViT-L/16 & 68.6 & 85.6 & 66.5 & 92.6 & 56.0 & 68.0 & 63.2 & 52.0 & 74.4 & 50.7 & 59.5 & 57.8 & 64.5 & 89.6 & 82.5 & 86.0 & 91.8 & 89.8 & 66.6 & 58.4 \\
ViT-L/16-IN21k & 68.1 & \bfseries 87.7 & 62.8 & 93.5 & 55.9 & 66.3 & 61.4 & 51.5 & 70.6 & 52.3 & 61.6 & \bfseries 60.3 & 65.0 & 87.5 & 80.5 & 84.4 & \bfseries 92.1 & 84.9 & 64.0 & 59.0 \\
ViT-B/16 & 67.9 & 81.1 & 65.1 & 92.8 & 55.5 & 69.0 & 64.1 & 50.5 & 68.3 & 54.5 & 58.0 & 55.5 & 64.7 & 90.8 & 84.3 & 86.3 & 89.1 & 84.2 & 65.2 & 58.9 \\
ViT-B/16-IN21k & 66.7 & 82.6 & 61.7 & 92.5 & 55.5 & 67.1 & 62.5 & 50.4 & 64.9 & 53.9 & 58.9 & 54.2 & 63.1 & 87.6 & 80.8 & 84.1 & 87.3 & 81.0 & 64.2 & 59.4 \\
ResNet-101 & 66.4 & 67.0 & 68.2 & 92.5 & 53.8 & 63.0 & 65.6 & 53.1 & 76.8 & 51.6 & 56.5 & 55.1 & 65.6 & 93.2 & 85.8 & 86.9 & 71.9 & 80.4 & 64.5 & 55.5 \\
ResNet-50 & 68.0 & 69.9 & 69.7 & 92.6 & 54.7 & 64.6 & 65.6 & \bfseries 54.6 & 81.4 & 54.5 & 56.4 & 55.0 & 69.5 & \bfseries 94.0 & 86.5 & 86.9 & 72.5 & 86.7 & 67.6 & 56.4 \\
ResNet-18 & 68.2 & 69.1 & 69.9 & 92.2 & 55.4 & 67.1 & 65.5 & 52.7 & 82.0 & 54.2 & 60.3 & 56.2 & 65.8 & 92.7 & 83.8 & 84.2 & 71.6 & 87.0 & 67.8 & 60.2 \\
ResNet-50 (finetuned) & 67.0 & 68.0 & 64.7 & 91.0 & 54.7 & 63.4 & 67.5 & 52.8 & 78.8 & \bfseries 54.7 & 59.1 & 52.0 & 67.2 & 92.9 & 83.9 & 83.4 & 73.4 & 89.2 & 64.8 & 56.0 \\
ResNet-18 (finetuned) & 68.0 & 69.4 & 70.2 & 91.3 & 55.7 & 66.4 & 68.8 & 53.0 & 81.5 & 52.6 & 57.5 & 52.7 & 66.1 & 92.8 & 83.9 & 82.4 & 75.9 & 90.2 & \bfseries 68.7 & 58.2 \\
\bottomrule
\end{tabular}

}
\\
\centering
\scalebox{.33}{

\begin{tabular}{lSSSSSSSSSSSSSSSSSSSS}
\toprule
{} & \multicolumn{20}{c}{$n=3$} \\
{} & {Harmonic Mean} & {aml} & {bus} & {crc} & {cxr} & {derm} & {dr\_regular} & {dr\_uwf} & {fundus} & {glaucoma} & {mammo\_calc} & {mammo\_mass} & {oct} & {organs\_axial} & {organs\_coronal} & {organs\_sagittal} & {pbc} & {pneumonia} & {skinl\_derm} & {skinl\_photo} \\

\midrule
BiomedCLIP-ViT-B/16 & \bfseries 73.2 & 80.8 & 78.6 & \bfseries 97.8 & \bfseries 61.4 & \bfseries 78.0 & 70.0 & 51.8 & \bfseries 83.8 & 52.0 & 64.2 & 57.3 & \bfseries 75.8 & \bfseries 95.3 & \bfseries 91.2 & \bfseries 90.1 & 92.9 & \bfseries 96.9 & \bfseries 70.9 & 61.6 \\
CLIP-ViT-g/14-laion2B & 72.6 & 79.0 & \bfseries 79.7 & 97.0 & 57.1 & 73.4 & \bfseries 72.1 & 54.1 & 81.6 & 54.2 & 65.0 & 59.6 & 72.3 & 94.7 & 89.8 & 88.8 & 87.4 & 93.6 & 70.5 & 59.7 \\
CLIP-ViT-H/14-laion2B & 73.0 & 80.9 & 76.8 & 97.2 & 56.5 & 73.5 & \bfseries 72.1 & 55.1 & 83.1 & 54.3 & \bfseries 66.6 & 60.2 & 74.7 & 93.4 & 88.2 & 88.8 & 88.7 & 92.5 & \bfseries 70.9 & 61.2 \\
CLIP-ViT-L/14 & 71.5 & 79.9 & 73.3 & 97.1 & 56.7 & 72.3 & 71.6 & 54.2 & 81.6 & 53.6 & 62.8 & 57.3 & 71.2 & 92.3 & 88.3 & 88.1 & 88.7 & 91.4 & 66.8 & 61.1 \\
CLIP-ViT-B/16 & 69.6 & 75.4 & 70.9 & 95.7 & 55.3 & 69.8 & 67.0 & 51.9 & 75.7 & 54.0 & 59.7 & 57.3 & 70.6 & 92.1 & 87.0 & 86.6 & 82.0 & 90.2 & 68.9 & 59.3 \\
DINOv2-ViT-g/14 & 71.2 & 80.1 & 67.2 & 97.4 & 56.9 & 72.8 & 69.3 & 52.1 & 79.1 & 52.9 & 59.2 & 60.6 & 70.7 & 93.7 & 86.5 & 89.4 & 85.9 & 95.1 & 70.3 & \bfseries 64.6 \\
DINOv2-ViT-L/14 & 70.2 & 78.6 & 64.3 & 97.0 & 56.7 & 71.4 & 72.0 & 51.0 & 76.8 & 54.4 & 57.2 & 58.6 & 70.3 & 92.7 & 86.2 & 88.7 & 85.4 & 94.3 & 68.2 & 61.6 \\
DINOv2-ViT-B/14 & 70.6 & 77.0 & 68.4 & 96.7 & 57.4 & 71.4 & 69.9 & 53.7 & 76.2 & 53.2 & 58.9 & 58.8 & 70.1 & 93.6 & 87.6 & 89.0 & 84.7 & 92.8 & 69.7 & 60.6 \\
ViT-H/14-IN21k & 71.4 & 85.9 & 68.8 & 95.8 & 56.7 & 73.5 & 67.6 & 50.7 & 76.3 & 55.9 & 61.7 & 60.4 & 72.1 & 92.7 & 87.2 & 89.7 & 93.8 & 89.8 & 69.6 & 60.9 \\
ViT-L/16 & 70.9 & 88.3 & 68.5 & 94.9 & 56.4 & 72.2 & 64.4 & 52.3 & 75.9 & 54.7 & 60.9 & 61.3 & 69.6 & 92.1 & 85.5 & 88.8 & 93.9 & 90.9 & 69.1 & 59.3 \\
ViT-L/16-IN21k & 70.3 & \bfseries 90.1 & 66.1 & 95.5 & 56.4 & 70.3 & 62.2 & 51.6 & 72.1 & 54.1 & 63.2 & \bfseries 62.4 & 70.3 & 90.3 & 84.0 & 87.8 & \bfseries 94.0 & 88.0 & 68.0 & 60.6 \\
ViT-B/16 & 70.5 & 84.2 & 68.2 & 95.0 & 56.7 & 72.1 & 65.5 & 51.3 & 70.1 & 56.2 & 62.2 & 59.3 & 70.2 & 93.2 & 87.1 & 89.2 & 91.8 & 89.7 & 66.7 & 61.0 \\
ViT-B/16-IN21k & 69.1 & 85.0 & 64.1 & 94.7 & 56.6 & 70.8 & 63.4 & 50.4 & 67.6 & 56.4 & 61.2 & 56.4 & 69.0 & 89.8 & 84.4 & 87.5 & 89.0 & 85.5 & 65.4 & 62.4 \\
ResNet-101 & 68.4 & 70.4 & 70.1 & 94.3 & 55.6 & 64.6 & 68.2 & 53.7 & 76.2 & 53.6 & 56.7 & 58.4 & 69.8 & 94.8 & 88.0 & 88.9 & 76.9 & 82.0 & 65.3 & 58.0 \\
ResNet-50 & 70.2 & 72.6 & 72.0 & 94.1 & 54.2 & 66.6 & 67.5 & \bfseries 56.2 & 82.0 & \bfseries 58.5 & 58.8 & 57.9 & 73.2 & 95.1 & 88.3 & 89.1 & 76.3 & 89.4 & 68.3 & 58.5 \\
ResNet-18 & 70.6 & 71.6 & 71.4 & 93.9 & 56.7 & 70.3 & 67.1 & 54.1 & 82.1 & 57.4 & 60.7 & 57.8 & 71.3 & 94.4 & 86.7 & 86.5 & 76.0 & 91.5 & 70.6 & 62.8 \\
ResNet-50 (finetuned) & 69.5 & 71.8 & 67.2 & 92.7 & 55.9 & 65.5 & 69.1 & 52.9 & 79.1 & 53.5 & 61.8 & 58.4 & 72.0 & 94.7 & 86.8 & 86.5 & 81.2 & 93.4 & 67.8 & 57.6 \\
ResNet-18 (finetuned) & 70.4 & 73.4 & 70.8 & 92.3 & 56.5 & 69.2 & 70.7 & 53.1 & 80.5 & 55.0 & 60.1 & 56.6 & 72.4 & 93.8 & 87.6 & 85.5 & 81.8 & 94.3 & 69.7 & 60.6 \\
\bottomrule
\end{tabular}

}
\\
\centering
\scalebox{.33}{

\begin{tabular}{lSSSSSSSSSSSSSSSSSSSS}
\toprule
{} & \multicolumn{20}{c}{$n=5$} \\
{} & {Harmonic Mean} & {aml} & {bus} & {crc} & {cxr} & {derm} & {dr\_regular} & {dr\_uwf} & {fundus} & {glaucoma} & {mammo\_calc} & {mammo\_mass} & {oct} & {organs\_axial} & {organs\_coronal} & {organs\_sagittal} & {pbc} & {pneumonia} & {skinl\_derm} & {skinl\_photo} \\

\midrule
BiomedCLIP-ViT-B/16 & \bfseries 76.1 & 85.1 & 83.4 & \bfseries 98.5 & \bfseries 63.1 & \bfseries 81.6 & 74.1 & 53.4 & \bfseries 86.6 & 55.6 & 64.7 & 60.9 & 79.5 & 96.6 & 93.5 & \bfseries 92.4 & 95.0 & \bfseries 96.5 & 74.8 & 64.6 \\
CLIP-ViT-g/14-laion2B & 75.5 & 84.4 & \bfseries 83.9 & 98.2 & 58.3 & 78.7 & \bfseries 75.8 & 55.0 & 84.8 & 54.8 & \bfseries 66.6 & 62.5 & 78.9 & 96.5 & 92.6 & 91.8 & 91.7 & 94.4 & 75.2 & 63.0 \\
CLIP-ViT-H/14-laion2B & 75.6 & 85.8 & 81.4 & 98.1 & 57.9 & 78.5 & 75.7 & 54.7 & 86.1 & 56.0 & 66.4 & 61.5 & \bfseries 79.6 & 95.7 & 91.3 & 91.2 & 93.2 & 94.6 & \bfseries 75.3 & 64.8 \\
CLIP-ViT-L/14 & 74.1 & 84.4 & 77.5 & 98.0 & 57.8 & 76.9 & 75.5 & 54.3 & 84.9 & 52.6 & 65.8 & 60.1 & 77.1 & 94.5 & 90.7 & 90.5 & 92.6 & 93.5 & 71.1 & 64.2 \\
CLIP-ViT-B/16 & 72.2 & 80.9 & 74.4 & 97.1 & 56.3 & 74.4 & 70.6 & 51.8 & 82.6 & 53.1 & 64.3 & 57.6 & 74.9 & 94.2 & 89.9 & 89.6 & 86.9 & 92.4 & 73.3 & 60.7 \\
DINOv2-ViT-g/14 & 74.3 & 85.5 & 72.3 & 98.4 & 58.2 & 77.8 & 72.7 & 53.9 & 83.0 & 54.2 & 62.2 & 63.1 & 75.7 & 95.7 & 90.6 & 91.9 & 91.1 & 95.3 & 74.3 & \bfseries 68.0 \\
DINOv2-ViT-L/14 & 73.4 & 83.4 & 68.9 & 98.0 & 57.8 & 76.4 & 74.5 & 52.5 & 82.4 & 57.0 & 60.2 & 61.8 & 75.2 & 94.6 & 89.3 & 90.7 & 89.7 & 94.9 & 72.9 & 65.7 \\
DINOv2-ViT-B/14 & 73.8 & 81.4 & 71.2 & 98.0 & 58.9 & 76.3 & 74.0 & 55.1 & 82.4 & 54.7 & 61.5 & 64.1 & 75.5 & 95.5 & 90.8 & 91.8 & 88.7 & 94.2 & 74.3 & 63.2 \\
ViT-H/14-IN21k & 74.4 & 90.3 & 75.7 & 97.4 & 58.1 & 77.4 & 70.9 & 51.1 & 80.7 & 59.0 & 63.5 & 62.4 & 75.6 & 95.2 & 90.6 & \bfseries 92.4 & \bfseries 96.4 & 91.9 & 73.2 & 64.7 \\
ViT-L/16 & 73.8 & 91.3 & 74.7 & 96.5 & 58.2 & 76.8 & 68.8 & 54.0 & 78.6 & 55.8 & 63.3 & 64.7 & 73.6 & 94.6 & 89.1 & 91.3 & 95.7 & 93.2 & 73.5 & 60.9 \\
ViT-L/16-IN21k & 73.4 & \bfseries 92.5 & 72.9 & 96.8 & 58.2 & 75.6 & 66.8 & 52.3 & 75.8 & 55.5 & 66.4 & \bfseries 65.8 & 73.5 & 93.3 & 88.1 & 90.5 & 95.9 & 90.4 & 72.9 & 62.5 \\
ViT-B/16 & 73.1 & 88.1 & 73.6 & 96.4 & 58.2 & 76.4 & 69.1 & 52.0 & 74.7 & \bfseries 59.2 & 61.5 & 60.5 & 73.8 & 95.0 & 90.1 & 91.5 & 94.6 & 90.4 & 70.9 & 64.3 \\
ViT-B/16-IN21k & 72.1 & 88.7 & 70.4 & 96.5 & 58.1 & 75.1 & 67.9 & 52.3 & 72.1 & 58.1 & 62.7 & 57.7 & 72.3 & 93.5 & 88.4 & 90.6 & 92.9 & 88.3 & 69.1 & 64.5 \\
ResNet-101 & 71.4 & 74.7 & 76.9 & 95.8 & 56.0 & 69.9 & 71.7 & 54.9 & 81.5 & 53.1 & 60.2 & 60.5 & 74.7 & 96.2 & 91.1 & 91.4 & 80.8 & 86.4 & 70.6 & 58.8 \\
ResNet-50 & 72.7 & 76.8 & 77.3 & 95.7 & 56.0 & 71.0 & 71.2 & \bfseries 57.6 & 85.0 & 57.9 & 60.2 & 58.4 & 78.2 & 96.5 & 91.2 & 91.4 & 80.3 & 90.8 & 72.8 & 60.3 \\
ResNet-18 & 73.3 & 76.9 & 78.4 & 95.6 & 57.4 & 73.2 & 71.6 & 55.0 & 84.8 & 57.9 & 65.5 & 58.6 & 75.1 & 96.3 & 90.3 & 90.2 & 81.2 & 93.7 & 73.5 & 63.2 \\
ResNet-50 (finetuned) & 73.0 & 79.6 & 74.4 & 95.7 & 53.1 & 70.1 & 73.4 & 55.5 & 83.1 & 57.9 & 64.9 & 57.6 & 78.6 & \bfseries 97.2 & 93.5 & 92.3 & 89.6 & 93.3 & 71.2 & 60.4 \\
ResNet-18 (finetuned) & 73.9 & 83.8 & 78.8 & 95.4 & 57.3 & 75.5 & 72.8 & 53.8 & 80.7 & 58.9 & 61.5 & 55.9 & 79.1 & 96.8 & \bfseries 93.7 & 92.0 & 91.5 & 92.2 & 74.6 & 63.6 \\
\bottomrule
\end{tabular}

}
\\
\centering
\scalebox{.33}{

\begin{tabular}{lSSSSSSSSSSSSSSSSSSSS}
\toprule
{} & \multicolumn{20}{c}{$n=7$} \\
{} & {Harmonic Mean} & {aml} & {bus} & {crc} & {cxr} & {derm} & {dr\_regular} & {dr\_uwf} & {fundus} & {glaucoma} & {mammo\_calc} & {mammo\_mass} & {oct} & {organs\_axial} & {organs\_coronal} & {organs\_sagittal} & {pbc} & {pneumonia} & {skinl\_derm} & {skinl\_photo} \\

\midrule
BiomedCLIP-ViT-B/16 & 77.4 & 88.0 & 84.9 & \bfseries 99.0 & \bfseries 65.1 & \bfseries 83.8 & 75.5 & 52.4 & 87.3 & 55.2 & 66.3 & 62.1 & 83.2 & 97.3 & 95.4 & \bfseries 93.8 & 96.5 & \bfseries 97.1 & 75.7 & 68.4 \\
CLIP-ViT-g/14-laion2B & 77.3 & 87.4 & \bfseries 86.3 & 98.5 & 60.1 & 80.9 & 78.3 & 54.1 & 86.6 & 55.9 & 68.4 & 63.3 & 81.7 & 97.0 & 94.0 & 93.1 & 93.8 & 96.3 & 76.8 & 68.2 \\
CLIP-ViT-H/14-laion2B & \bfseries 77.5 & 88.2 & 83.4 & 98.4 & 59.9 & 81.0 & \bfseries 79.0 & 55.0 & \bfseries 88.2 & 57.7 & \bfseries 68.9 & 62.9 & 82.5 & 96.4 & 93.1 & 92.9 & 94.8 & 95.4 & 76.3 & 68.7 \\
CLIP-ViT-L/14 & 76.3 & 88.2 & 79.7 & 98.4 & 59.2 & 79.0 & 76.9 & 55.0 & 85.1 & 57.7 & 67.0 & 61.2 & 80.8 & 95.6 & 92.4 & 91.4 & 95.0 & 95.5 & 73.1 & 68.7 \\
CLIP-ViT-B/16 & 74.2 & 84.0 & 76.9 & 97.8 & 57.6 & 76.5 & 73.8 & 51.8 & 82.6 & 55.6 & 64.0 & 59.9 & 79.2 & 95.3 & 91.6 & 91.0 & 90.8 & 93.9 & 75.0 & 65.8 \\
DINOv2-ViT-g/14 & 76.4 & 88.5 & 75.9 & 98.7 & 59.9 & 79.7 & 75.0 & 53.5 & 83.9 & 55.4 & 65.7 & 64.0 & 81.1 & 96.7 & 92.7 & 93.2 & 93.5 & 96.6 & \bfseries 76.9 & \bfseries 72.5 \\
DINOv2-ViT-L/14 & 75.5 & 86.7 & 71.7 & 98.5 & 59.7 & 78.6 & 77.1 & 52.8 & 83.8 & 59.4 & 61.6 & 61.8 & 80.0 & 95.9 & 92.0 & 92.6 & 92.9 & 95.9 & 75.8 & 69.9 \\
DINOv2-ViT-B/14 & 75.4 & 85.4 & 75.0 & 98.4 & 60.4 & 78.2 & 75.5 & 54.7 & 82.2 & 55.4 & 62.0 & 62.9 & 79.2 & 96.5 & 92.7 & 93.1 & 92.2 & 95.5 & 76.7 & 67.9 \\
ViT-H/14-IN21k & 75.5 & 91.4 & 77.3 & 97.9 & 59.6 & 80.1 & 73.8 & 50.7 & 82.4 & 56.4 & 65.2 & 62.3 & 78.7 & 96.2 & 92.6 & 93.6 & 96.6 & 93.5 & 75.8 & 67.0 \\
ViT-L/16 & 75.4 & 92.7 & 77.1 & 97.3 & 59.6 & 79.2 & 71.4 & 52.3 & 83.0 & 57.7 & 65.4 & 64.8 & 76.6 & 95.8 & 91.0 & 92.8 & \bfseries 96.8 & 94.0 & 74.6 & 63.9 \\
ViT-L/16-IN21k & 75.1 & \bfseries 93.3 & 74.0 & 97.4 & 59.2 & 78.4 & 70.3 & 51.8 & 80.8 & 56.9 & 66.9 & \bfseries 65.7 & 77.0 & 94.6 & 90.3 & 92.0 & 96.7 & 92.5 & 74.1 & 66.6 \\
ViT-B/16 & 74.6 & 89.1 & 75.0 & 97.1 & 58.9 & 79.4 & 72.6 & 51.0 & 77.9 & 59.3 & 64.6 & 59.9 & 76.6 & 95.9 & 91.9 & 93.0 & 95.6 & 92.6 & 72.0 & 67.6 \\
ViT-B/16-IN21k & 73.5 & 89.7 & 71.2 & 97.0 & 59.0 & 77.9 & 71.1 & 51.6 & 76.0 & 58.6 & 62.7 & 57.9 & 75.3 & 94.6 & 90.2 & 91.9 & 94.2 & 91.4 & 70.3 & 67.9 \\
ResNet-101 & 73.1 & 79.1 & 77.6 & 96.8 & 58.2 & 72.1 & 74.2 & \bfseries 56.1 & 83.9 & 55.1 & 60.8 & 60.4 & 76.7 & 97.0 & 92.5 & 92.4 & 85.6 & 88.5 & 69.4 & 62.0 \\
ResNet-50 & 74.4 & 80.2 & 79.2 & 96.7 & 58.7 & 72.8 & 73.7 & 55.9 & 85.7 & 61.3 & 60.1 & 58.2 & 80.1 & 97.2 & 92.8 & 92.6 & 85.1 & 92.9 & 73.2 & 64.7 \\
ResNet-18 & 75.3 & 79.5 & 80.0 & 96.5 & 59.1 & 76.0 & 74.0 & 54.3 & 87.0 & 60.2 & 68.1 & 59.2 & 77.6 & 97.1 & 92.5 & 92.0 & 84.0 & 95.6 & 75.1 & 68.0 \\
ResNet-50 (finetuned) & 74.4 & 84.0 & 74.3 & 95.5 & 58.7 & 72.1 & 74.8 & 55.0 & 83.7 & 59.5 & 64.6 & 58.7 & 80.8 & 96.9 & 93.7 & 91.7 & 93.4 & 93.8 & 71.6 & 62.0 \\
ResNet-18 (finetuned) & 76.5 & 88.1 & 82.5 & 96.6 & 59.0 & 78.8 & 76.0 & 53.3 & 84.8 & \bfseries 62.7 & 63.8 & 58.1 & \bfseries 83.9 & \bfseries 97.9 & \bfseries 95.7 & 93.5 & 94.9 & 94.5 & 75.7 & 67.8 \\
\bottomrule
\end{tabular}

}
\\
\centering
\scalebox{.33}{

\begin{tabular}{lSSSSSSSSSSSSSSSSSSSS}
\toprule
{} & \multicolumn{20}{c}{$n=10$} \\
{} & {Harmonic Mean} & {aml} & {bus} & {crc} & {cxr} & {derm} & {dr\_regular} & {dr\_uwf} & {fundus} & {glaucoma} & {mammo\_calc} & {mammo\_mass} & {oct} & {organs\_axial} & {organs\_coronal} & {organs\_sagittal} & {pbc} & {pneumonia} & {skinl\_derm} & {skinl\_photo} \\

\midrule
BiomedCLIP-ViT-B/16 & \bfseries 79.1 & 91.3 & 85.1 & \bfseries 99.0 & \bfseries 65.9 & \bfseries 84.9 & 75.9 & 53.4 & 88.7 & 60.1 & 69.0 & 63.7 & 84.9 & 97.9 & 96.3 & \bfseries 94.9 & 97.3 & \bfseries 98.2 & 78.2 & 70.3 \\
CLIP-ViT-g/14-laion2B & 78.8 & 89.9 & \bfseries 87.7 & 98.7 & 60.5 & 83.6 & 79.5 & 56.6 & 87.4 & 57.1 & 70.1 & 64.4 & 84.3 & 97.8 & 95.0 & 94.3 & 95.6 & 96.5 & \bfseries 79.1 & 70.3 \\
CLIP-ViT-H/14-laion2B & \bfseries 79.1 & 90.5 & 86.4 & 98.7 & 60.7 & 83.1 & \bfseries 79.6 & 57.7 & \bfseries 88.9 & 58.6 & \bfseries 70.5 & 63.4 & 85.3 & 97.3 & 94.4 & 93.8 & 96.6 & 96.8 & 78.9 & 71.3 \\
CLIP-ViT-L/14 & 77.8 & 89.7 & 83.3 & 98.6 & 59.8 & 82.2 & 78.9 & 56.6 & 87.5 & 55.4 & 69.5 & 63.6 & 82.8 & 96.4 & 93.7 & 93.1 & 95.4 & 95.6 & 75.9 & 70.8 \\
CLIP-ViT-B/16 & 75.6 & 86.5 & 80.1 & 97.9 & 58.5 & 79.4 & 74.5 & 54.0 & 86.7 & 53.3 & 65.2 & 61.3 & 80.9 & 96.2 & 93.0 & 92.6 & 92.9 & 93.9 & 77.2 & 67.3 \\
DINOv2-ViT-g/14 & 78.3 & 89.5 & 79.9 & 98.8 & 60.3 & 82.3 & 77.6 & 55.8 & 86.6 & 54.8 & 68.6 & \bfseries 68.5 & 83.3 & 97.3 & 94.2 & 94.3 & 94.3 & 96.4 & 79.0 & \bfseries 75.4 \\
DINOv2-ViT-L/14 & 77.5 & 90.0 & 76.9 & 98.8 & 60.2 & 81.0 & 77.9 & 55.7 & 84.9 & 58.9 & 62.4 & 67.3 & 82.5 & 96.9 & 93.7 & 93.5 & 95.6 & 96.3 & 77.0 & 71.8 \\
DINOv2-ViT-B/14 & 77.7 & 88.6 & 79.7 & 98.7 & 60.9 & 80.6 & 77.1 & 56.6 & 84.4 & 56.2 & 66.2 & 67.1 & 82.0 & 97.3 & 94.1 & 93.9 & 95.2 & 95.5 & 78.8 & 71.0 \\
ViT-H/14-IN21k & 77.4 & 93.8 & 80.5 & 98.3 & 60.2 & 82.4 & 74.4 & 51.7 & 82.8 & 60.3 & 66.8 & 65.5 & 81.7 & 97.0 & 94.2 & 94.2 & \bfseries 97.8 & 94.8 & 78.0 & 68.8 \\
ViT-L/16 & 76.8 & 92.9 & 81.0 & 97.5 & 60.0 & 81.2 & 72.1 & 55.0 & 83.3 & 55.8 & 66.5 & 67.7 & 79.6 & 96.1 & 92.2 & 93.6 & 96.5 & 95.8 & 77.1 & 67.1 \\
ViT-L/16-IN21k & 76.8 & \bfseries 94.2 & 77.7 & 97.9 & 59.8 & 80.6 & 70.9 & 53.7 & 82.4 & 57.7 & 68.3 & 68.1 & 79.8 & 95.4 & 92.0 & 93.1 & 96.9 & 94.3 & 76.7 & 69.2 \\
ViT-B/16 & 76.1 & 91.6 & 79.2 & 97.5 & 60.3 & 80.7 & 73.4 & 52.3 & 79.8 & 59.3 & 63.7 & 62.4 & 78.6 & 96.8 & 93.3 & 93.1 & 97.0 & 94.7 & 74.3 & 69.6 \\
ViT-B/16-IN21k & 75.6 & 92.1 & 76.0 & 97.5 & 60.3 & 79.8 & 72.8 & 53.6 & 77.3 & 58.7 & 64.3 & 60.9 & 77.5 & 96.0 & 92.2 & 92.8 & 96.1 & 94.7 & 72.9 & 70.7 \\
ResNet-101 & 75.0 & 80.8 & 81.7 & 97.3 & 58.2 & 74.1 & 75.6 & 57.1 & 85.4 & 56.9 & 64.0 & 61.4 & 79.7 & 97.7 & 93.8 & 93.8 & 87.6 & 90.6 & 73.3 & 64.3 \\
ResNet-50 & 76.6 & 82.8 & 81.5 & 97.4 & 59.2 & 75.3 & 74.4 & \bfseries 59.0 & 86.8 & 62.8 & 64.6 & 60.9 & 82.2 & 98.1 & 94.6 & 93.9 & 88.4 & 94.0 & 75.3 & 68.2 \\
ResNet-18 & 77.0 & 81.3 & 81.9 & 97.2 & 59.6 & 78.2 & 75.2 & 55.7 & 87.9 & \bfseries 63.0 & 68.1 & 62.9 & 80.0 & 97.6 & 93.8 & 93.2 & 86.7 & 96.1 & 77.4 & 70.0 \\
ResNet-50 (finetuned) & 76.7 & 87.8 & 78.0 & 96.5 & 59.7 & 74.2 & 76.6 & 58.7 & 86.1 & 61.7 & 65.5 & 61.5 & 83.1 & 98.1 & 95.2 & 93.4 & 95.3 & 94.4 & 73.6 & 64.8 \\
ResNet-18 (finetuned) & 78.6 & 90.4 & 84.7 & 97.4 & 61.1 & 80.8 & 77.9 & 57.7 & 87.4 & 62.1 & 66.9 & 61.6 & \bfseries 86.1 & \bfseries 98.3 & \bfseries 96.7 & 94.8 & 95.6 & 95.2 & 78.2 & 69.5 \\
\bottomrule
\end{tabular}

}
\\
\centering
\scalebox{.33}{

\begin{tabular}{lSSSSSSSSSSSSSSSSSSSS}
\toprule
{} & \multicolumn{20}{c}{$n=15$} \\
{} & {Harmonic Mean} & {aml} & {bus} & {crc} & {cxr} & {derm} & {dr\_regular} & {dr\_uwf} & {fundus} & {glaucoma} & {mammo\_calc} & {mammo\_mass} & {oct} & {organs\_axial} & {organs\_coronal} & {organs\_sagittal} & {pbc} & {pneumonia} & {skinl\_derm} & {skinl\_photo} \\

\midrule
BiomedCLIP-ViT-B/16 & 80.3 & 92.5 & 87.0 & 99.2 & \bfseries 67.4 & \bfseries 87.5 & 78.2 & 54.1 & 88.2 & 61.1 & 67.4 & 65.3 & 87.8 & 98.2 & 96.9 & 95.8 & 97.9 & \bfseries 98.0 & 79.8 & 73.3 \\
CLIP-ViT-g/14-laion2B & 80.8 & 92.8 & \bfseries 88.5 & 99.2 & 62.0 & 87.0 & \bfseries 82.5 & 57.6 & 87.2 & 62.0 & \bfseries 69.5 & 65.6 & 88.9 & 98.4 & 96.7 & 95.6 & 97.2 & 97.0 & 81.9 & 73.5 \\
CLIP-ViT-H/14-laion2B & \bfseries 80.9 & 92.7 & 87.3 & 99.1 & 61.5 & 86.1 & 82.0 & 58.0 & 89.0 & 65.1 & 69.3 & 64.3 & 88.8 & 98.0 & 95.8 & 95.2 & 98.0 & 97.4 & \bfseries 82.0 & 74.1 \\
CLIP-ViT-L/14 & 79.9 & 92.2 & 84.5 & 98.9 & 60.8 & 84.8 & 81.5 & 57.9 & 88.3 & 61.7 & 69.2 & 64.4 & 86.6 & 97.2 & 94.9 & 94.6 & 97.3 & 96.3 & 79.2 & 73.4 \\
CLIP-ViT-B/16 & 77.3 & 89.2 & 80.7 & 98.5 & 60.0 & 82.3 & 78.0 & 54.2 & 86.4 & 56.9 & 67.3 & 60.2 & 85.2 & 97.0 & 94.9 & 93.9 & 95.3 & 94.6 & 78.4 & 70.7 \\
DINOv2-ViT-g/14 & 80.1 & 93.2 & 82.0 & \bfseries 99.3 & 61.9 & 85.4 & 80.6 & 56.1 & 85.8 & 61.5 & 67.0 & 68.8 & 87.0 & 98.2 & 95.6 & 95.5 & 97.1 & 96.8 & 81.6 & \bfseries 76.2 \\
DINOv2-ViT-L/14 & 79.1 & 92.3 & 79.0 & 99.0 & 61.4 & 84.4 & 81.2 & 54.7 & 85.7 & 64.4 & 63.0 & 66.2 & 86.0 & 97.5 & 95.0 & 94.9 & 96.7 & 95.9 & 80.3 & 74.1 \\
DINOv2-ViT-B/14 & 79.3 & 91.1 & 81.6 & 99.0 & 62.0 & 83.6 & 79.9 & 56.5 & 85.0 & 61.3 & 65.0 & 65.6 & 86.4 & 97.9 & 95.5 & 95.4 & 96.4 & 96.5 & 81.5 & 74.3 \\
ViT-H/14-IN21k & 79.3 & 93.4 & 82.8 & 98.6 & 61.2 & 85.5 & 77.2 & 53.0 & 85.8 & 62.8 & 68.0 & 67.5 & 85.8 & 97.5 & 95.3 & 95.5 & 97.8 & 96.2 & 79.8 & 73.7 \\
ViT-L/16 & 79.1 & 94.8 & 83.3 & 98.3 & 60.8 & 85.1 & 75.7 & 56.3 & 85.5 & 60.7 & 66.0 & \bfseries 70.1 & 84.0 & 97.2 & 94.1 & 95.1 & 97.8 & 95.5 & 80.1 & 70.7 \\
ViT-L/16-IN21k & 78.9 & \bfseries 96.0 & 82.0 & 98.6 & 61.5 & 84.6 & 74.5 & 54.1 & 83.6 & 61.9 & 66.1 & 69.2 & 84.0 & 97.0 & 93.9 & 94.5 & \bfseries 98.5 & 95.8 & 79.5 & 72.2 \\
ViT-B/16 & 78.2 & 91.5 & 81.1 & 98.0 & 60.9 & 83.8 & 76.3 & 52.0 & 82.3 & 64.4 & 67.3 & 64.0 & 83.5 & 97.3 & 94.6 & 95.0 & 97.3 & 95.2 & 78.1 & 72.4 \\
ViT-B/16-IN21k & 77.7 & 92.7 & 79.0 & 98.2 & 61.1 & 83.3 & 75.0 & 54.1 & 80.4 & 63.0 & 66.0 & 62.4 & 83.3 & 96.9 & 93.8 & 94.4 & 96.8 & 95.1 & 76.4 & 72.4 \\
ResNet-101 & 77.2 & 84.9 & 82.9 & 98.0 & 59.3 & 77.5 & 78.4 & 59.1 & 86.7 & 61.1 & 64.1 & 62.6 & 83.4 & 98.1 & 95.0 & 94.9 & 91.2 & 92.6 & 75.7 & 67.3 \\
ResNet-50 & 78.1 & 85.1 & 82.5 & 97.9 & 60.0 & 78.4 & 77.5 & 58.4 & 88.3 & 66.9 & 63.7 & 60.7 & 86.5 & 98.3 & 95.4 & 95.1 & 90.5 & 95.3 & 78.2 & 71.0 \\
ResNet-18 & 78.5 & 84.8 & 83.4 & 97.7 & 60.2 & 81.1 & 77.1 & 57.0 & \bfseries 89.1 & 65.0 & 67.5 & 63.2 & 83.6 & 98.1 & 95.0 & 94.5 & 90.3 & 97.0 & 78.7 & 72.2 \\
ResNet-50 (finetuned) & 78.5 & 90.4 & 81.0 & 97.7 & 60.2 & 78.6 & 79.2 & \bfseries 59.6 & 86.5 & 64.6 & 65.1 & 59.9 & 88.0 & 98.8 & 96.7 & 94.8 & 97.1 & 97.1 & 78.1 & 68.5 \\
ResNet-18 (finetuned) & 79.7 & 93.9 & 87.4 & 97.9 & 60.8 & 84.6 & 79.1 & 56.1 & 88.7 & \bfseries 67.2 & 64.4 & 58.0 & \bfseries 91.0 & \bfseries 99.0 & \bfseries 98.0 & \bfseries 96.0 & 97.8 & 96.8 & 80.4 & 72.5 \\
\bottomrule
\end{tabular}

}
\\
\centering
\scalebox{.33}{

\begin{tabular}{lSSSSSSSSSSSSSSSSSSSS}
\toprule
{} & \multicolumn{20}{c}{$n=20$} \\
{} & {Harmonic Mean} & {aml} & {bus} & {crc} & {cxr} & {derm} & {dr\_regular} & {dr\_uwf} & {fundus} & {glaucoma} & {mammo\_calc} & {mammo\_mass} & {oct} & {organs\_axial} & {organs\_coronal} & {organs\_sagittal} & {pbc} & {pneumonia} & {skinl\_derm} & {skinl\_photo} \\

\midrule
BiomedCLIP-ViT-B/16 & 81.9 & 93.7 & 88.9 & 99.3 & \bfseries 68.2 & 88.5 & 79.2 & 56.6 & 89.5 & 63.1 & 71.3 & 67.3 & 89.6 & 98.6 & 97.7 & 96.3 & 98.2 & \bfseries 98.3 & 82.9 & 74.7 \\
CLIP-ViT-g/14-laion2B & 82.3 & 93.9 & \bfseries 91.2 & 99.2 & 63.0 & \bfseries 88.7 & 82.4 & 60.6 & 89.7 & 61.3 & \bfseries 73.3 & 67.8 & 90.1 & 98.8 & 97.4 & 96.1 & 97.9 & 97.7 & 84.7 & 75.3 \\
CLIP-ViT-H/14-laion2B & \bfseries 82.5 & 93.7 & 89.9 & 99.2 & 62.9 & 88.1 & \bfseries 82.7 & 61.3 & \bfseries 91.4 & 64.7 & 72.4 & 65.9 & 90.2 & 98.4 & 96.6 & 95.8 & 98.3 & 97.8 & 84.3 & 76.5 \\
CLIP-ViT-L/14 & 81.2 & 93.4 & 87.8 & 99.1 & 61.9 & 86.2 & 82.0 & 60.8 & 89.8 & 60.8 & 70.8 & 66.9 & 87.8 & 97.7 & 95.8 & 94.9 & 97.6 & 97.3 & 80.8 & 76.0 \\
CLIP-ViT-B/16 & 79.0 & 90.5 & 85.1 & 98.8 & 60.9 & 84.7 & 78.9 & 55.6 & 89.6 & 57.5 & 69.3 & 61.6 & 86.7 & 97.4 & 95.7 & 94.7 & 95.7 & 96.0 & 82.0 & 72.8 \\
DINOv2-ViT-g/14 & 82.2 & 94.2 & 86.3 & \bfseries 99.4 & 63.0 & 86.9 & 81.7 & 58.9 & 88.6 & 63.8 & 71.0 & 68.9 & 89.6 & 98.6 & 96.6 & 96.1 & 97.6 & 97.8 & \bfseries 85.2 & \bfseries 79.4 \\
DINOv2-ViT-L/14 & 81.0 & 93.5 & 83.6 & 99.2 & 62.5 & 86.0 & 81.4 & 58.4 & 88.2 & 64.9 & 66.3 & 68.3 & 87.7 & 98.1 & 95.8 & 95.2 & 97.3 & 97.5 & 82.6 & 76.0 \\
DINOv2-ViT-B/14 & 81.2 & 92.0 & 85.5 & 99.2 & 62.9 & 85.3 & 81.3 & 59.1 & 87.7 & 63.3 & 68.7 & 67.1 & 87.8 & 98.4 & 96.3 & 96.1 & 96.9 & 97.3 & 84.8 & 76.1 \\
ViT-H/14-IN21k & 80.3 & 95.4 & 85.1 & 99.1 & 62.2 & 87.1 & 78.0 & 53.0 & 86.6 & 63.6 & 70.4 & 67.1 & 86.4 & 98.4 & 96.7 & 96.1 & \bfseries 98.6 & 97.0 & 82.0 & 73.3 \\
ViT-L/16 & 80.2 & 96.0 & 85.4 & 98.8 & 62.0 & 86.2 & 76.9 & 56.8 & 87.1 & 61.1 & 68.8 & 69.3 & 85.5 & 97.9 & 95.4 & 95.5 & 98.4 & 96.6 & 82.2 & 71.0 \\
ViT-L/16-IN21k & 80.3 & \bfseries 96.4 & 84.3 & 98.8 & 62.4 & 86.3 & 75.6 & 56.2 & 86.0 & 61.1 & 70.6 & \bfseries 70.8 & 86.2 & 97.6 & 95.1 & 95.2 & 98.5 & 96.8 & 80.9 & 73.4 \\
ViT-B/16 & 79.5 & 93.9 & 84.7 & 98.6 & 62.3 & 85.6 & 77.3 & 54.0 & 83.7 & 63.4 & 67.9 & 65.4 & 84.9 & 98.1 & 95.8 & 95.2 & 98.1 & 97.1 & 79.6 & 73.9 \\
ViT-B/16-IN21k & 79.3 & 94.3 & 82.0 & 98.6 & 62.6 & 85.0 & 76.9 & 55.8 & 83.1 & 64.1 & 68.9 & 64.0 & 84.3 & 97.6 & 95.1 & 94.8 & 97.7 & 96.8 & 77.9 & 73.6 \\
ResNet-101 & 78.6 & 85.7 & 85.6 & 98.2 & 60.0 & 80.4 & 79.5 & 61.5 & 89.1 & 60.4 & 65.9 & 63.1 & 84.8 & 98.5 & 95.8 & 95.5 & 91.5 & 94.3 & 78.8 & 69.5 \\
ResNet-50 & 79.4 & 86.7 & 84.9 & 98.3 & 61.4 & 80.8 & 78.0 & 59.5 & 89.0 & 65.3 & 65.6 & 62.1 & 87.6 & 98.8 & 96.5 & 95.6 & 92.8 & 95.6 & 80.9 & 73.4 \\
ResNet-18 & 80.0 & 85.9 & 86.2 & 98.1 & 61.7 & 82.8 & 78.3 & 58.5 & 90.1 & 65.6 & 70.5 & 64.5 & 84.9 & 98.5 & 96.1 & 95.4 & 91.4 & 97.6 & 81.5 & 73.4 \\
ResNet-50 (finetuned) & 80.8 & 92.7 & 83.6 & 98.3 & 61.1 & 80.2 & 80.4 & 62.3 & 88.6 & \bfseries 67.8 & 70.7 & 63.8 & 89.2 & 99.1 & 97.8 & 95.9 & 97.7 & 97.8 & 80.4 & 69.9 \\
ResNet-18 (finetuned) & 82.4 & 94.2 & 89.9 & 98.7 & 62.9 & 86.4 & 81.6 & \bfseries 62.7 & 90.3 & 67.2 & 72.0 & 63.6 & \bfseries 91.8 & \bfseries 99.2 & \bfseries 98.4 & \bfseries 96.4 & 97.8 & 97.5 & 83.6 & 74.3 \\
\bottomrule
\end{tabular}

}
\\
\centering
\scalebox{.33}{

\begin{tabular}{lSSSSSSSSSSSSSSSSSSSS}
\toprule
{} & \multicolumn{20}{c}{$n=25$} \\
{} & {Harmonic Mean} & {aml} & {bus} & {crc} & {cxr} & {derm} & {dr\_regular} & {dr\_uwf} & {fundus} & {glaucoma} & {mammo\_calc} & {mammo\_mass} & {oct} & {organs\_axial} & {organs\_coronal} & {organs\_sagittal} & {pbc} & {pneumonia} & {skinl\_derm} & {skinl\_photo} \\

\midrule
BiomedCLIP-ViT-B/16 & 82.6 & 94.2 & 89.2 & 99.4 & \bfseries 68.8 & 89.0 & 79.9 & 56.4 & 90.0 & 65.2 & 71.4 & 69.3 & 91.2 & 98.8 & 98.0 & 96.6 & 98.6 & \bfseries 99.0 & 82.5 & 75.7 \\
CLIP-ViT-g/14-laion2B & 83.4 & 94.3 & \bfseries 91.9 & 99.4 & 63.7 & \bfseries 89.9 & 83.4 & 62.2 & 90.0 & 63.0 & \bfseries 74.9 & 69.9 & 92.2 & 98.9 & 97.6 & 96.5 & 98.0 & 98.1 & 84.0 & 77.9 \\
CLIP-ViT-H/14-laion2B & \bfseries 83.8 & 94.5 & 90.8 & 99.4 & 63.4 & 89.1 & 83.2 & \bfseries 63.8 & \bfseries 91.2 & 67.7 & 74.4 & 69.0 & 92.0 & 98.6 & 97.2 & 96.0 & \bfseries 98.8 & 98.4 & 83.2 & 79.4 \\
CLIP-ViT-L/14 & 82.4 & 94.4 & 88.7 & 99.2 & 63.1 & 87.5 & 82.9 & 61.7 & 89.8 & 62.8 & 73.5 & 68.1 & 90.1 & 98.1 & 96.6 & 95.4 & 98.2 & 97.9 & 80.9 & 78.6 \\
CLIP-ViT-B/16 & 80.2 & 91.8 & 86.2 & 98.9 & 62.0 & 86.3 & 79.7 & 55.5 & 89.1 & 60.1 & 71.2 & 65.6 & 88.6 & 97.8 & 96.3 & 95.2 & 96.7 & 96.0 & 81.2 & 74.7 \\
DINOv2-ViT-g/14 & 83.1 & 95.2 & 87.0 & \bfseries 99.5 & 63.9 & 88.1 & 82.9 & 58.1 & 88.6 & 65.2 & 72.4 & \bfseries 72.7 & 91.4 & 99.0 & 97.1 & 96.3 & 98.4 & 98.1 & 84.6 & \bfseries 81.1 \\
DINOv2-ViT-L/14 & 82.1 & 94.5 & 85.0 & 99.3 & 62.8 & 87.5 & \bfseries 83.7 & 57.4 & 88.3 & 66.3 & 69.4 & 70.4 & 90.5 & 98.5 & 96.5 & 95.9 & 98.1 & 97.6 & 82.8 & 78.6 \\
DINOv2-ViT-B/14 & 82.3 & 93.2 & 86.6 & 99.4 & 63.5 & 86.9 & 82.4 & 58.9 & 87.9 & 64.8 & 69.9 & 71.1 & 90.2 & 98.6 & 96.7 & 96.4 & 97.7 & 97.5 & \bfseries 84.7 & 77.9 \\
ViT-H/14-IN21k & 81.4 & 95.4 & 86.8 & 99.2 & 62.4 & 88.3 & 79.4 & 53.6 & 86.6 & 65.9 & 71.7 & 69.3 & 89.2 & 98.6 & 97.0 & 96.4 & 98.6 & 97.1 & 82.6 & 76.2 \\
ViT-L/16 & 81.3 & 96.1 & 86.8 & 98.9 & 62.3 & 87.2 & 77.9 & 57.3 & 87.4 & 63.2 & 71.4 & 71.4 & 87.5 & 98.0 & 95.5 & 96.1 & 98.4 & 97.3 & 82.1 & 73.9 \\
ViT-L/16-IN21k & 81.1 & \bfseries 96.8 & 85.8 & 99.0 & 63.0 & 87.3 & 77.4 & 55.7 & 87.1 & 64.5 & 70.0 & 71.0 & 87.8 & 97.9 & 95.6 & 95.7 & \bfseries 98.8 & 96.7 & 81.4 & 74.9 \\
ViT-B/16 & 80.3 & 95.0 & 85.5 & 98.8 & 62.5 & 86.2 & 78.1 & 53.9 & 84.5 & 65.7 & 68.9 & 67.6 & 86.6 & 98.5 & 96.3 & 95.6 & 98.6 & 96.9 & 79.5 & 75.1 \\
ViT-B/16-IN21k & 80.1 & 94.2 & 83.9 & 98.7 & 62.0 & 85.9 & 77.6 & 56.2 & 83.8 & 64.6 & 70.6 & 65.7 & 86.0 & 97.8 & 95.5 & 95.4 & 97.6 & 96.5 & 78.8 & 75.1 \\
ResNet-101 & 79.7 & 87.9 & 87.6 & 98.6 & 61.5 & 81.1 & 80.5 & 61.0 & 89.1 & 63.2 & 68.4 & 64.1 & 87.2 & 98.8 & 96.4 & 95.8 & 93.8 & 94.5 & 77.6 & 70.5 \\
ResNet-50 & 80.3 & 87.5 & 87.2 & 98.4 & 61.7 & 81.9 & 79.1 & 60.4 & 89.5 & 67.0 & 68.4 & 62.0 & 89.5 & 98.9 & 96.8 & 95.9 & 92.8 & 96.4 & 80.5 & 74.2 \\
ResNet-18 & 80.9 & 87.6 & 88.0 & 98.4 & 62.1 & 83.8 & 79.4 & 58.1 & 90.2 & 68.4 & 73.0 & 64.1 & 88.0 & 98.7 & 96.5 & 95.8 & 92.2 & 97.4 & 81.3 & 74.4 \\
ResNet-50 (finetuned) & 81.7 & 94.1 & 85.8 & 98.5 & 61.1 & 80.9 & 82.3 & 63.2 & 88.5 & 70.1 & 69.6 & 67.1 & 91.4 & 99.3 & 98.2 & 96.3 & 98.5 & 97.7 & 79.7 & 71.6 \\
ResNet-18 (finetuned) & 83.3 & 95.5 & 90.9 & 98.9 & 65.0 & 87.8 & 82.8 & 60.0 & 90.5 & \bfseries 71.1 & 72.2 & 66.7 & \bfseries 94.2 & \bfseries 99.4 & \bfseries 98.6 & \bfseries 96.8 & 98.4 & 98.0 & 82.8 & 75.2 \\
\bottomrule
\end{tabular}

}
\\
\centering
\scalebox{.33}{

\begin{tabular}{lSSSSSSSSSSSSSSSSSSSS}
\toprule
{} & \multicolumn{20}{c}{$n=30$} \\
{} & {Harmonic Mean} & {aml} & {bus} & {crc} & {cxr} & {derm} & {dr\_regular} & {dr\_uwf} & {fundus} & {glaucoma} & {mammo\_calc} & {mammo\_mass} & {oct} & {organs\_axial} & {organs\_coronal} & {organs\_sagittal} & {pbc} & {pneumonia} & {skinl\_derm} & {skinl\_photo} \\

\midrule
BiomedCLIP-ViT-B/16 & 83.6 & 94.8 & 89.5 & \bfseries 99.4 & \bfseries 69.9 & 89.7 & 80.5 & 57.7 & 89.5 & 69.7 & 72.5 & 70.5 & 91.4 & 98.8 & 98.1 & 96.7 & 98.4 & \bfseries 98.6 & 83.3 & 77.6 \\
CLIP-ViT-g/14-laion2B & 84.3 & 95.5 & \bfseries 91.5 & \bfseries 99.4 & 64.3 & \bfseries 90.5 & \bfseries 84.7 & 61.5 & 89.9 & 67.4 & \bfseries 75.2 & 71.9 & 92.5 & 99.0 & 97.8 & 96.7 & 98.4 & 97.9 & 86.0 & 79.4 \\
CLIP-ViT-H/14-laion2B & \bfseries 84.6 & 95.1 & 91.1 & 99.3 & 64.5 & 89.7 & 84.3 & 63.3 & \bfseries 91.1 & 71.2 & 74.9 & 70.9 & 91.7 & 98.8 & 97.3 & 96.3 & \bfseries 98.7 & 98.5 & 85.4 & 80.5 \\
CLIP-ViT-L/14 & 83.3 & 95.3 & 88.8 & 99.3 & 63.4 & 88.2 & 83.4 & 61.9 & 89.8 & 66.7 & 73.1 & 70.3 & 90.8 & 98.5 & 96.8 & 95.5 & 98.5 & 97.7 & 83.0 & 79.5 \\
CLIP-ViT-B/16 & 81.0 & 92.9 & 86.4 & 99.0 & 62.8 & 86.8 & 81.6 & 55.8 & 89.2 & 61.3 & 72.8 & 66.6 & 88.6 & 98.1 & 96.6 & 95.4 & 97.2 & 96.4 & 83.3 & 75.9 \\
DINOv2-ViT-g/14 & 84.0 & 95.7 & 87.6 & \bfseries 99.4 & 64.8 & 88.6 & 84.2 & 59.3 & 89.0 & 68.5 & 72.8 & \bfseries 73.2 & 92.0 & 99.0 & 97.3 & 96.6 & 98.2 & 98.0 & \bfseries 86.4 & \bfseries 83.0 \\
DINOv2-ViT-L/14 & 83.1 & 94.8 & 85.0 & 99.3 & 63.9 & 88.0 & 84.2 & 59.8 & 89.3 & 68.8 & 69.9 & 71.9 & 90.9 & 98.5 & 96.6 & 96.1 & 98.0 & 97.3 & 85.1 & 80.3 \\
DINOv2-ViT-B/14 & 83.1 & 93.8 & 86.9 & 99.3 & 64.5 & 87.1 & 82.8 & 61.1 & 88.1 & 67.5 & 70.4 & 72.0 & 90.0 & 98.7 & 97.0 & 96.3 & 97.5 & 97.1 & 85.3 & 80.2 \\
ViT-H/14-IN21k & 82.1 & 95.3 & 87.3 & 99.1 & 63.3 & 88.9 & 79.8 & 53.2 & 88.0 & 70.0 & 74.4 & 69.5 & 88.8 & 98.6 & 97.0 & 96.4 & 98.4 & 96.2 & 83.3 & 77.7 \\
ViT-L/16 & 81.8 & 96.2 & 87.0 & 98.9 & 63.1 & 88.1 & 78.8 & 56.8 & 88.0 & 66.3 & 71.5 & 71.5 & 87.7 & 98.4 & 96.0 & 96.2 & 98.3 & 96.5 & 83.1 & 74.8 \\
ViT-L/16-IN21k & 82.0 & \bfseries 96.9 & 86.8 & 99.0 & 63.8 & 88.0 & 78.2 & 56.1 & 87.1 & 67.8 & 71.3 & 72.4 & 88.3 & 98.1 & 95.9 & 95.8 & \bfseries 98.7 & 96.7 & 82.9 & 76.2 \\
ViT-B/16 & 81.5 & 94.5 & 86.4 & 98.7 & 63.1 & 87.4 & 79.6 & 55.4 & 85.5 & 68.4 & 71.1 & 70.2 & 86.8 & 98.4 & 96.6 & 95.8 & 98.3 & 95.8 & 81.5 & 77.1 \\
ViT-B/16-IN21k & 80.9 & 95.2 & 84.4 & 98.8 & 63.4 & 86.8 & 78.9 & 55.5 & 84.7 & 68.5 & 70.2 & 68.3 & 85.9 & 98.2 & 96.1 & 95.4 & 98.2 & 95.8 & 79.5 & 76.0 \\
ResNet-101 & 80.2 & 88.6 & 87.4 & 98.6 & 61.8 & 81.5 & 81.2 & 60.3 & 89.3 & 63.6 & 68.5 & 65.7 & 87.8 & 98.8 & 96.5 & 95.9 & 94.1 & 94.7 & 79.4 & 72.0 \\
ResNet-50 & 81.0 & 89.4 & 87.4 & 98.6 & 63.0 & 82.5 & 79.8 & 58.8 & 88.7 & 70.5 & 68.0 & 62.6 & 89.9 & 99.0 & 97.1 & 96.1 & 94.8 & 96.8 & 82.2 & 75.8 \\
ResNet-18 & 81.5 & 87.7 & 87.0 & 98.3 & 62.8 & 84.1 & 80.0 & 59.5 & 90.7 & 70.4 & 74.7 & 65.0 & 86.8 & 98.7 & 96.7 & 95.9 & 91.9 & 97.4 & 81.9 & 75.8 \\
ResNet-50 (finetuned) & 82.6 & 93.9 & 87.0 & 98.8 & 61.9 & 84.1 & 82.0 & \bfseries 63.6 & 88.6 & 72.9 & 71.0 & 65.9 & 92.0 & \bfseries 99.4 & 98.1 & 96.1 & 98.5 & 97.6 & 82.3 & 74.8 \\
ResNet-18 (finetuned) & 84.1 & 95.6 & 90.8 & 98.9 & 64.6 & 88.5 & 83.3 & 62.8 & 91.0 & \bfseries 73.8 & 74.3 & 67.1 & \bfseries 93.7 & \bfseries 99.4 & \bfseries 98.6 & \bfseries 96.8 & 98.3 & 98.0 & 83.7 & 76.7 \\
\bottomrule
\end{tabular}

}

%% file: bibliography.bib
@misc{woerner2024comprehensive,
	title={A comprehensive and easy-to-use multi-domain multi-task medical imaging meta-dataset (MedIMeta)},
	author={Stefano Woerner and Arthur Jaques and Christian F. Baumgartner},
	year={2024},
	eprint={2404.16000},
	archivePrefix={arXiv},
	primaryClass={cs.CV}
}

@dataset{stefano_woerner_2024_7884735,
	author       = {Stefano Woerner and
	Arthur Jaques and
	Christian Baumgartner},
	title        = {{MedIMeta: A comprehensive and easy-to-use multi- 
	domain multi-task medical imaging meta-dataset}},
	month        = apr,
	year         = 2024,
	publisher    = {Zenodo},
	doi          = {10.5281/zenodo.7884735},
	url          = {https://doi.org/10.5281/zenodo.7884735}
}

@inproceedings{
	woerner2022strategies,
	title={Strategies for Meta-Learning with Diverse Tasks},
	author={Stefano Woerner and Christian F. Baumgartner},
	booktitle={Medical Imaging with Deep Learning},
	year={2022}
}

@article{ILSVRC15,
Author = {Olga Russakovsky and Jia Deng and Hao Su and Jonathan Krause and Sanjeev Satheesh and Sean Ma and Zhiheng Huang and Andrej Karpathy and Aditya Khosla and Michael Bernstein and Alexander C. Berg and Li Fei-Fei},
Title = { {ImageNet Large Scale Visual Recognition Challenge} },
Year = {2015},
journal   = {International Journal of Computer Vision (IJCV)},
doi = {10.1007/s11263-015-0816-y},
volume={115},
number={3},
pages={211-252}
}

@misc{ridnik2021imagenet21k,
title={ImageNet-21K Pretraining for the Masses},
author={Tal Ridnik and Emanuel Ben-Baruch and Asaf Noy and Lihi Zelnik-Manor},
year={2021},
eprint={2104.10972},
archivePrefix={arXiv},
primaryClass={cs.CV}
}

@InProceedings{He_2016_CVPR,
	author = {He, Kaiming and Zhang, Xiangyu and Ren, Shaoqing and Sun, Jian},
	title = {Deep Residual Learning for Image Recognition},
	booktitle = {Proceedings of the IEEE Conference on Computer Vision and Pattern Recognition (CVPR)},
	month = {June},
	year = {2016}
}

@InProceedings{pmlr-v139-radford21a,
	title = 	 {Learning Transferable Visual Models From Natural Language Supervision},
	author =       {Radford, Alec and Kim, Jong Wook and Hallacy, Chris and Ramesh, Aditya and Goh, Gabriel and Agarwal, Sandhini and Sastry, Girish and Askell, Amanda and Mishkin, Pamela and Clark, Jack and Krueger, Gretchen and Sutskever, Ilya},
	booktitle = 	 {Proceedings of the 38th International Conference on Machine Learning},
	pages = 	 {8748--8763},
	year = 	 {2021},
	editor = 	 {Meila, Marina and Zhang, Tong},
	volume = 	 {139},
	series = 	 {Proceedings of Machine Learning Research},
	month = 	 {18--24 Jul},
	publisher =    {PMLR},
	url = 	 {https://proceedings.mlr.press/v139/radford21a.html}
}

@misc{dosovitskiy2020image,
	title={An Image is Worth 16x16 Words: Transformers for Image Recognition at Scale},
	author={Alexey Dosovitskiy and Lucas Beyer and Alexander Kolesnikov and Dirk Weissenborn and Xiaohua Zhai and Thomas Unterthiner and Mostafa Dehghani and Matthias Minderer and Georg Heigold and Sylvain Gelly and Jakob Uszkoreit and Neil Houlsby},
	year={2020},
	eprint={2010.11929},
	archivePrefix={arXiv},
	primaryClass={cs.CV}
}

@misc{zhang2023biomedclip,
	title={BiomedCLIP: a multimodal biomedical foundation model pretrained from fifteen million scientific image-text pairs},
	author={Sheng Zhang and Yanbo Xu and Naoto Usuyama and Hanwen Xu and Jaspreet Bagga and Robert Tinn and Sam Preston and Rajesh Rao and Mu Wei and Naveen Valluri and Cliff Wong and Andrea Tupini and Yu Wang and Matt Mazzola and Swadheen Shukla and Lars Liden and Jianfeng Gao and Matthew P. Lungren and Tristan Naumann and Sheng Wang and Hoifung Poon},
	year={2023},
	eprint={2303.00915},
	archivePrefix={arXiv},
	primaryClass={cs.CV}
}

@article{Gu_2021,
   title={Domain-Specific Language Model Pretraining for Biomedical Natural Language Processing},
   volume={3},
   ISSN={2637-8051},
   url={http://dx.doi.org/10.1145/3458754},
   DOI={10.1145/3458754},
   number={1},
   journal={ACM Transactions on Computing for Healthcare},
   publisher={Association for Computing Machinery (ACM)},
   author={Gu, Yu and Tinn, Robert and Cheng, Hao and Lucas, Michael and Usuyama, Naoto and Liu, Xiaodong and Naumann, Tristan and Gao, Jianfeng and Poon, Hoifung},
   year={2021},
   month=oct, pages={1–23} }

@inproceedings{caron2021emerging,
  title={Emerging properties in self-supervised vision transformers},
  author={Caron, Mathilde and Touvron, Hugo and Misra, Ishan and J{\'e}gou, Herv{\'e} and Mairal, Julien and Bojanowski, Piotr and Joulin, Armand},
  booktitle={Proceedings of the IEEE/CVF international conference on computer vision},
  pages={9650--9660},
  year={2021}
}

@article{oquab2023dinov2,
  title={Dinov2: Learning robust visual features without supervision},
  author={Oquab, Maxime and Darcet, Timoth{\'e}e and Moutakanni, Th{\'e}o and Vo, Huy and Szafraniec, Marc and Khalidov, Vasil and Fernandez, Pierre and Haziza, Daniel and Massa, Francisco and El-Nouby, Alaaeldin and others},
  journal={arXiv preprint arXiv:2304.07193},
  year={2023}
}

@misc{kingma2017adam,
      title={Adam: A Method for Stochastic Optimization}, 
      author={Diederik P. Kingma and Jimmy Ba},
      year={2017},
      eprint={1412.6980},
      archivePrefix={arXiv},
      primaryClass={cs.LG}
}

@misc{schuhmann2022laion5b,
    title={LAION-5B: An open large-scale dataset for training next generation image-text models},
    author={Christoph Schuhmann and Romain Beaumont and Richard Vencu and Cade Gordon and Ross Wightman and Mehdi Cherti and Theo Coombes and Aarush Katta and Clayton Mullis and Mitchell Wortsman and Patrick Schramowski and Srivatsa Kundurthy and Katherine Crowson and Ludwig Schmidt and Robert Kaczmarczyk and Jenia Jitsev},
    year={2022},
    eprint={2210.08402},
    archivePrefix={arXiv},
    primaryClass={cs.CV}
}

@inproceedings{Cherti_2023,
   title={Reproducible Scaling Laws for Contrastive Language-Image Learning},
   url={http://dx.doi.org/10.1109/CVPR52729.2023.00276},
   DOI={10.1109/cvpr52729.2023.00276},
   booktitle={2023 IEEE/CVF Conference on Computer Vision and Pattern Recognition (CVPR)},
   publisher={IEEE},
   author={Cherti, Mehdi and Beaumont, Romain and Wightman, Ross and Wortsman, Mitchell and Ilharco, Gabriel and Gordon, Cade and Schuhmann, Christoph and Schmidt, Ludwig and Jitsev, Jenia},
   year={2023},
   month=jun
}

@misc{commoncrawl,
  title = {Common Crawl},
  howpublished = {\url{https://commoncrawl.org}}
}
